\documentclass{article}

\usepackage[english]{babel}

\usepackage[a4paper,top=2cm,bottom=2cm,left=3cm,right=3cm,marginparwidth=1.75cm]{geometry}

\usepackage{cite}
\usepackage{amsmath,amssymb,amsfonts}
\usepackage{algorithmic}
\usepackage{graphicx}
\usepackage{textcomp}
\usepackage{xcolor}
\usepackage{subcaption}
\usepackage{amsmath}
\usepackage{graphicx}
\usepackage[colorlinks=true, allcolors=blue]{hyperref}

\title{Multiresolution Neural Networks for Imaging}

\author{Hallison Paz, Tiago Novello, Vinicius Silva, Luiz Schirmer, \\ Guilherme Schardong, Fabio Chagas, Helio Lopes, Luiz Velho}

\date{}

\begin{document}
\maketitle

\begin{abstract}
We present MR-Net, a general architecture for multiresolution neural networks, and a framework for imaging applications based on this architecture. Our coordinate-based networks are continuous both in space and in scale as they are composed of multiple stages that progressively add finer details. Besides that, they are a compact and efficient representation. We show examples of multiresolution image representation and applications to texture magnification, minification, and antialiasing.

This document is the extended version of the paper~\cite{paz2022mrnet}. It includes additional material that would not fit the page limitations of the conference track for publication.
\end{abstract}

\section{Introduction}

Imaging applications benefit greatly from representations that support multiple resolutions. This is because they are instrumental for many tasks in computer vision and graphics, such as: compression; analysis and rendering. Traditionally, multiresolution representations for images have been based on signal processing techniques derived from Fourier theory.

Recently, the revolution in the media industry caused by deep neural networks motivated the development of new image representations adapted to machine learning methods.
In that context, we introduce a new framework for the representation of images in multiresolution using coordinate-based sinusoidal neural networks.

\subsection{Motivation}

The main breakthrough in deep learning for computer vision and imaging was due to the seminal work of LeCun, Bengio, and Hinton~\cite{cnn98}. This paper proposes the Convolutional Neural Network (CNN) as a proper architecture for the analysis of visual imagery.
The effectiveness of CNNs comes from the translation invariant properties of the convolution operator.

Deep multi-layer perceptron networks, such as CNNs, employ an array-based discrete representation of the underlying signal. In this case, the network input consists of a vector of pixel values (R,G,B) that represents the image \emph{directly} by data samples. Therefore, we can also call this kind of network a data-based network.

In contrast, a coordinate-based neural network represents the image \emph{indirectly}. This kind of network is a fully connected MLP (Multi Layer Perceptron) that takes as input a pixel coordinate $(x,y)$ and outputs the (R,G,B) color at that location.

While the data-based network is appropriate for analysis tasks, relying on a discrete description of the image, the coordinate-based network is suitable for synthesis, providing a continuous image description. For its characteristics, namely continuity and compactness, there is a growing interest in the research community to explore the potential of coordinate-based networks for imaging applications. 

\subsection{Related Work}

Coordinate-based networks provide a continuous functional description for images using an implicit neural representation~\cite{chen2021learning}. Since the coordinates are continuous, images can be presented in arbitrary resolution.

Spectral neural implicit architectures constitute a particular form of neural implicit representation in which the non-linear activation function is the periodic function $\sin(x)$. As such, it bridges the gap between the spatial and spectral domains, given the close relationship of the sine function with the Fourier~basis.

However, these neural representations with periodic activation functions have been regarded as difficult to train~\cite{taming2017}. To overcome this problem, Sitzmann et al.~\cite{sitzmann2019siren} proposed a sinusoidal network for signal representation called SIREN. One of the key contributions of this work is the initialization scheme that guarantees stability and good convergence. Furthermore, it also allows modeling fine details in accordance with the signal’s frequency content.

A multiplicative filter network (MFN)~\cite{fathony2020multiplicative} is a spectral neural implicit architecture simpler than SIREN which is equivalent to a shallow sinusoidal network. Lindell et al.~\cite{bacon2021} presented BACON (Band-limited Coordinate Network), an MFN that produces intermediate outputs with an analytical spectral bandwidth which can be specified at initialization and achieves multi-resolution decomposition of the output.

The control of frequency bands in the representation is closely related with the capability of adaptive reconstruction of the signal in multiple levels of detail.
In that context, Mueller et al.~\cite{mueller2022instant} developed a multiresolution neural network architecture based on hash encoding. Also, Martel et al.~\cite{martel2021acorn} designed an adaptive coordinate network for neural signal representation.

Another benefit of multiresolution image representations is the built-in support for antialiasing, which traditionally is implemented using image pyramids, such as in MIP Mapping~\cite{mipmap83}.

One of the important applications of imaging in both 2D and 3D Computer Graphics is texture synthesis.
In that realm, besides antialiasing, the creation of visual patterns from exemplars has great relevance~\cite{thies19}.
Spectral neural implicit architectures are particularly suited to model stationary or quasi-stationary signals due to the periodic nature of its activation function~\cite{chen2022}.

\subsection{Contributions}
In summary, we make the following contributions:
\begin{itemize}
  \item We introduce a family of multiresolution coordinate-based networks, with unified architecture, that provides a continuous representation spatially and in scale.
  \item We develop a framework for imaging applications based on this architecture, leveraging classical multiresolution concepts such as pyramids.
  \item We show that our architecture can represent images with good visual quality, outperforming related methods both in PSNR and number of parameters; we also demonstrate its use in applications of texture magnification and minification, and antialiasing. 
\end{itemize}

\section{Multiresolution Sinusoidal Neural Networks}

In this section we present \emph{MR-Net} (Multiresolution Sinusoidal Neural Networks), a representation of signals in multiple levels of detail using deep neural networks.


\subsection{Overview}

Our proposal is a family of coordinate-based networks with unified architecture. We derive three main variants, namely \emph{S-Net}, \emph{L-Net}, and \emph{M-Net}. As a whole, they provide different~trade-offs with respect to control of frequencies in the~representation. 

The characteristics of the MR-Net Family are:

\begin{itemize}
\item 2 Types of Level of Detail -- in this respect it can be based on network capacity or spectral projection.

\item 3 Types of Sampling -- the input signal can be given either by regular sampling (with or without subsampling) or by stochastic / stratified sampling.

\item Progressive Training -- the network is trained progressively using a variety of schedule regimes.

\item Continuous Multiscale -- the representation is continuous both in space and scale. Therefore it can reconstruct the signal at any desired resolution / level of detail.

\end{itemize}

The following subsections present the concepts involved in our approach, as well as, the technical details of the MR-Net architecture.
For a more complete description see~\cite{supplemental}.

\subsection{Architecture}

Based on considerations for learning and splitting frequencies of the input signal into levels of detail, we devised a general architecture for a family of neural networks.

The core idea is to structure the network into multiple \emph{stages}. Each stage learns in a controlled way a level of detail corresponding to a frequency band.

\begin{figure}[!h]
\centering
\includegraphics[width=0.99\linewidth]{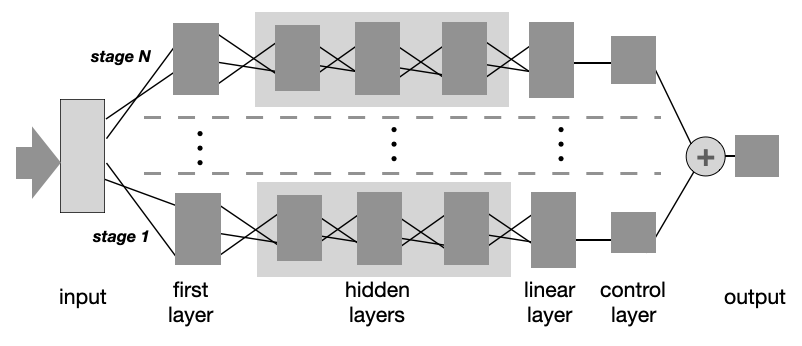}
\caption{Anatomy of MR-Net Family}
\label{f:mrnet-arch}
\end{figure}

A stage configuration is derived from a sub-network called MR-Module, which is
composed of four blocks of layers: the first layer; the hidden layers; the linear layer; and the control layer. 
The three initial blocks form a fully connected sinusoidal MLP. The last block consists of one node (not trained) to control the stage output, given by the function $c_\alpha({x}) = \alpha x$, with $\alpha \in [0,1]$, where ${x}$ comes from the linear layer. Figure~\ref{f:mrnet-arch} depicts the anatomy of the network, showing $N$ stages of MR-Modules. 

We can say that the first layer performs a projection of the input signal into a dictionary of sine functions, the hidden layers correspond to correlations of order $n$ of signal frequencies, the linear layer reconstructs the signal as a combination of these frequency atoms, and the control layer is just a mechanism to provide a continuous blend of level of detail in the network.

The stages of the network are trained based on a predefined schedule (See Section~\ref{s:training}).
During training, the control layer is the identity function, i.e. $\alpha = 1$.

The contribution of these $N$ stages is added together forming the network output.
Assuming that the MR-Net is learning a function $f(x)$ that fits the input signal, then
\begin{equation}
    f(x) = g_0(x) + \cdots + g_N(x)
\end{equation}
where $g_i(x)$ is the detail function given by the stage $s_i$, for $i = 1,\ldots,N$.
The first stage $s_0$ corresponds to the coarsest approximation of the signal and the other subsequent stages add increasingly finer details to it.

Note that this architecture is very much in the spirit of the Multiresolution Analysis~\cite{mallat-mr89}. Indeed, consider the base case with $f(x) = g_0(x) + g_1(x)$, then $g_1(x) = f(x) - g_0(x)$, i.e., $g_1$ are the details that need to be added to $g_0$ to increase the level of detail.

The network learns the decomposition of the signal as the projection into the coarse scale space and a sequence of finer detail spaces.
The characteristics of the level of detail decomposition of each member of the MR-Net family will depend on the specific configuration of the network stages, as will be presented next.

\subsubsection{S-Net}

\begin{figure}[!h]
\centering
\includegraphics[width=0.64\linewidth]{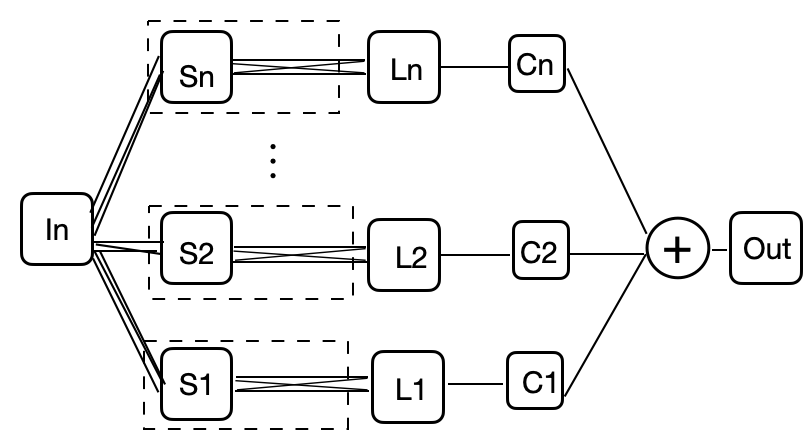}
\caption{S-Net}
\label{f:s-net}
\end{figure}

In the S-Net, a stage has only the first layer, the linear layer, and the control layer, which is not involved in the learning process. (See Figure~\ref{f:s-net}.)
Therefore, this kind of network consists of a learned ``sine transform", with these two layers corresponding, respectively, to the direct and inverse sine transform.
As a consequence, the S-Net can provide level of detail by the initialization of frequency bands in each stage.

\subsubsection{L-Net}

\begin{figure}[!h]
\centering
\includegraphics[width=0.60\linewidth]{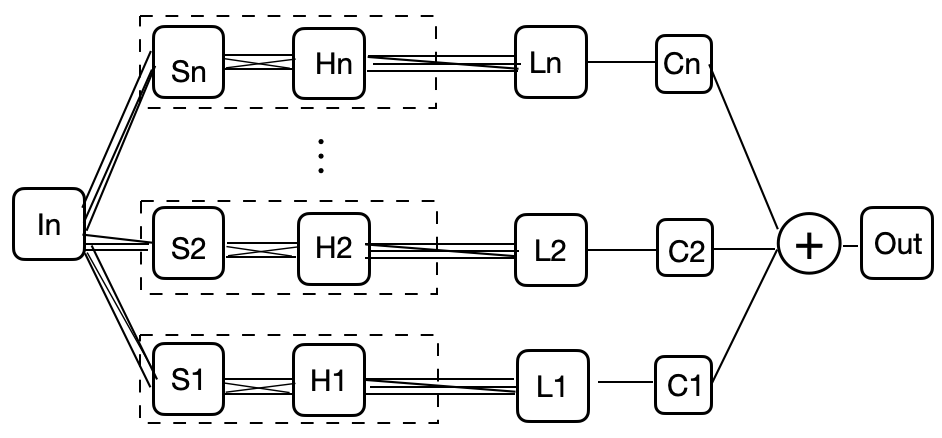}
\caption{L-Net}
\label{f:l-net}
\end{figure}

The L-Net is composed of $N$ complete independent stages (with the four blocks: first; hidden; linear and control) that are aggregated in the output
(see Figure~\ref{f:l-net}).
Consequently, the level of detail in this kind of network is determined by the capacity of the individual stages.

\subsubsection{M-Net}

\begin{figure}[!h]
\centering
\includegraphics[width=0.7\linewidth]{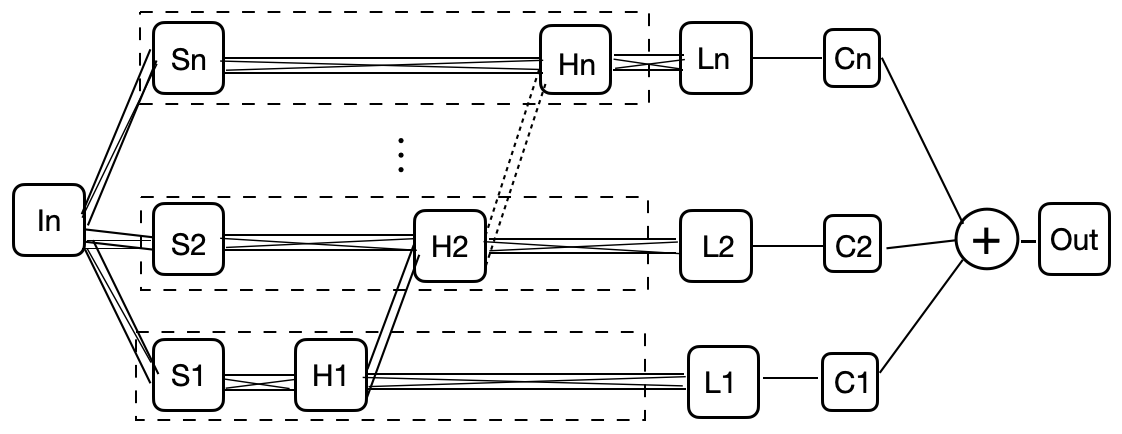}
\caption{M-Net}
\label{f:m-net}
\end{figure}

The M-Net consists of a hierarchy of stages, in which subsequent stages are linked together. The output of a block of hidden layers is connected both to the linear layer, as well as, to the input of the hidden layer block of the next stage (see Figure~\ref{f:m-net}).

A consequence of this hierarchical structure is that the hidden layer block in a stage is augmented with the sequence of hidden layer blocks coming from previous stages. Therefore, the capacity of each stage increases with its depth in the hierarchy.
Accordingly, it is expected that this kind of network provides a more powerful mechanism for learning levels of detail. For this reason we will use this variant for the imaging applications in this paper.

\subsection{Training}
\label{s:training}

The training of the MR-Net has to take into account the mechanisms for learning different levels of detail by each stage of the network. There are two basic mechanisms: i) level of detail filtering and ii) pre-processing of the input signal.

Level of detail filtering is achieved either with the frequency band filtering (through the initialization of the $w_0$ weights of the first layer); or with the capacity filtering (conditioned on the number of nodes and layers of the network).

Regarding the network input, we can either use the original signal, or pre-process the signal with a low-pass filter.

\subsubsection{Multi-Stage Schedule}

Since the M-Net has multiple stages, each learning a different level of detail, one important aspect is the stage training schedule. We could train the whole network with all stages in parallel, but we found it beneficial to train each stage in sequence from the lowest to the highest level of detail.
This scheduling is our choice and a common strategy in the traditional multiresolution analysis of signals.

\subsubsection{Progressive learning}

Furthermore, we adopt a progressive learning strategy by ``freezing" the weights of a stage once it is trained in the schedule sequence. This strategy guarantees that the details are added to the representation incrementally from coarse to fine.

\subsubsection{Adaptive Training}

We also employ an adaptive training scheme for the optimization of each network stage combining both accuracy loss thresholds and convergence rates.

\subsection{Level of Detail Schemes}

By incorporating the different aspects discussed in the previous sections we can define various schemes for learning level of detail representations using the family of Multiresolution Sinusoidal Neural Networks. The main ones are: Filtering with Gaussian Tower; and Filtering with Gaussian Pyramid. Here we highlight these two.

\subsubsection{Gaussian Tower}

If we want to have control over the frequencies present in the signal we can build a multiscale representation of the signal and train the network to approximate each level of this representation. 

We start by feeding our model with a ``Gaussian Tower", that is, multiple versions of the signal filtered consecutively by a low-pass filter, but without decimation. This way, each scale is reconstructed from the same amount of samples. The network must be trained from the less detailed scale to the most detailed one. 

\subsubsection{Gaussian Pyramid}

Based on the Shannon sampling theorem the Gaussian Tower is a highly redundant multiscale representation.
On the other hand, the Gaussian Pyramid is ”critically sampled”, i.e., it has the minimum number of samples required to represent each frequency band. 
The Gaussian Pyramid is a classical multiscale representation of uniformly sampled signals and we will adopt this scheme in the imaging applications of this paper.

\section{Imaging Applications}

In this section we describe the implementation and experiments of the MR-Net for imaging applications.
As already mentioned, we will adopt the M-Net variant of the architecture and a level of detail scheme based on the Gaussian Pyramid.

We designed the MR Module considering an empirical exploration of the sub-network capacity to represent images with typical characteristics (i.e., photographs). The configuration is as follows: width 96 neurons (fist, hidden and linear layers); number of layers of the hidden block equal to one.

The number of stages of the network is determined by the resolution of the image to be represented. The multiresolution for the image Pyramid is according to a dyadic structure, i.e., $2^j$. The base resolution of the first stage is $2^3 = 8$.

\begin{figure*}[!t]
\centering
\includegraphics[width=0.24\linewidth]{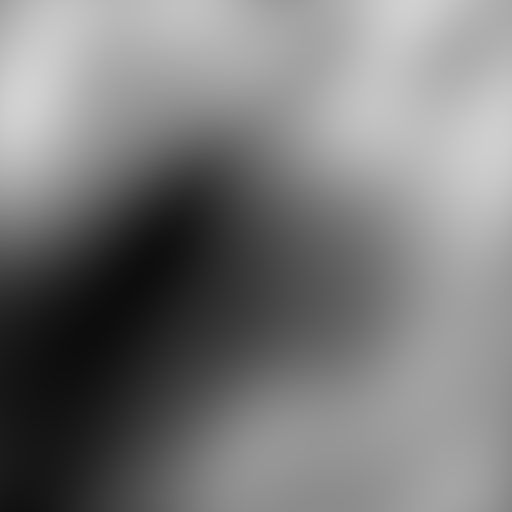}
\includegraphics[width=0.24\linewidth]{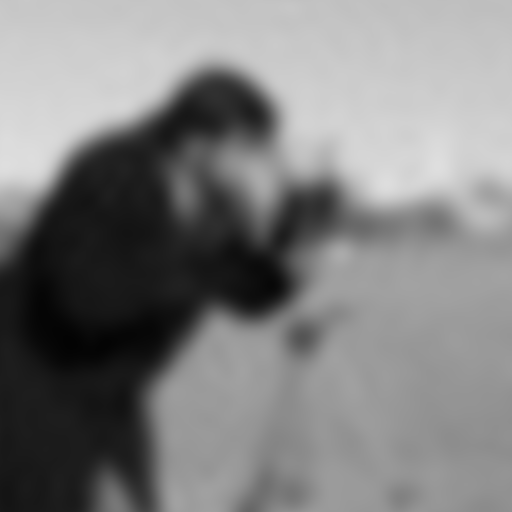}
\includegraphics[width=0.24\linewidth]{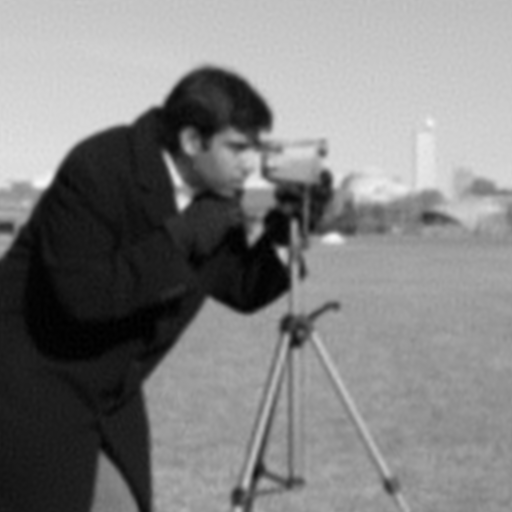}
\includegraphics[width=0.24\linewidth]{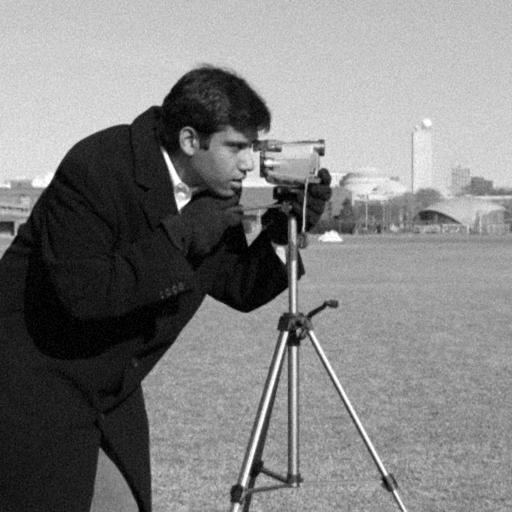} \\
\quad\vspace{-8pt} \\
\includegraphics[width=0.24\linewidth]{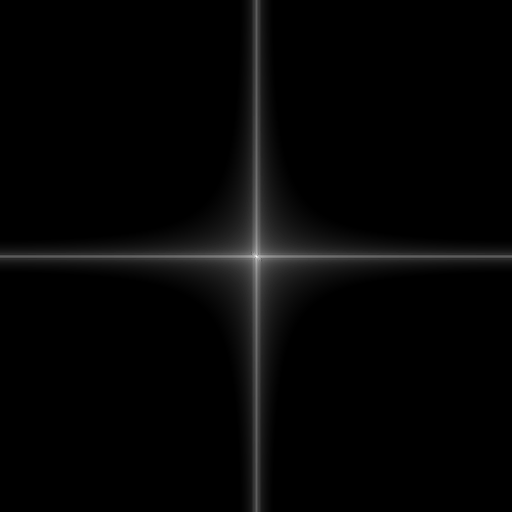}
\includegraphics[width=0.24\linewidth]{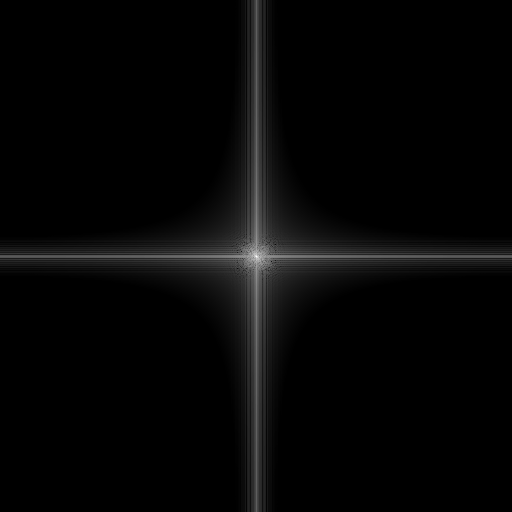}
\includegraphics[width=0.24\linewidth]{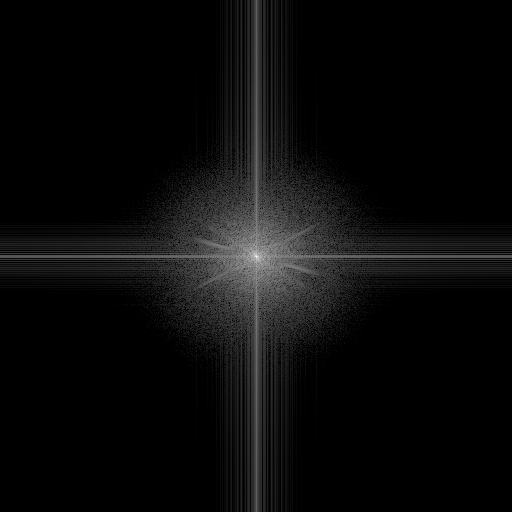}
\includegraphics[width=0.24\linewidth]{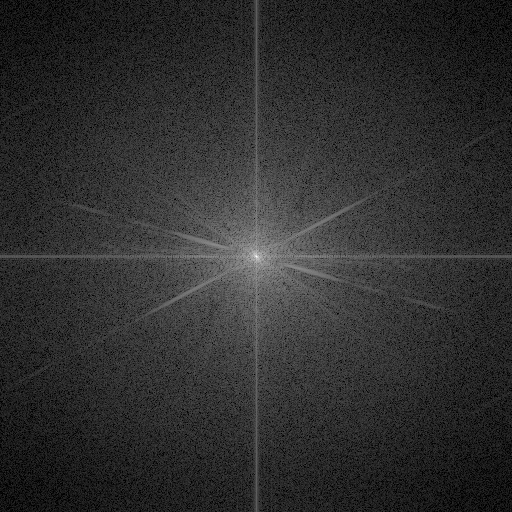}
\caption{Cameraman - reconstructed multiresolution levels 1, 3, 5 and 7 and corresponding Fourier spectra.}
\label{f:lod}
\end{figure*}

The Image Pyramid is built by filtering with a Gaussian kernel and decimation.
Most of the images used in the experiments have a resolution of 512. So, the pyramid is composed of the following resolution levels: 8, 16, 32, 64, 128, 256, and 512. Consequently, the network has a total of 7 stages.

We train the network using an adaptive scheme with the following hyper-parameters: Loss Function = MSE (Mean Squared Error); convergence threshold = $0.001$ (i.e., training of a stage stops if loss value changes less than $0.001$ percent); maximum number of epochs per stage = 300; each epoch visits all the pixels once; size of mini-batch = 65536 (to fit the GPU memory). Training is done with ADAM and learning rate of~$0.0001$.

The initialization of the network follows the scheme in~\cite{sitzmann2019siren}, that normalizes the weights of all layers and includes a factor $\omega_0$ that sets the spatial frequency of the first layer to better match the frequency spectrum of the signal. However, we differ in that the $\omega_0$ factor is used just for the initialization of the first layer and we create an additional factor, $\omega_G$ that is applied to the other layers to boost the network gradients. Also, the weights of first layer of each stage are set to match the frequencies of the corresponding level of detail, in the following way --- $\omega_0$ is uniformly distributed as $ \mathcal{U}(-B_i, B_i)$ with $B_i = [4, 8, 16, 32, 64, 128, 256]_{i=1}^7$ for a network with seven stages. The factor $\omega_G = 30$  for all experiments.

\subsection{Level of Detail Example}
\label{ss:LOD}

We now show an example of a multiresolution image representation using the setup described above. For this experiment we chose the "Cameraman", a standard test image used in the field of image processing and also in~\cite{sitzmann2019siren}.
The source is a monochromatic picture with $512\times 512$ pixels of resolution.

Figure~\ref{f:lod} depicts the levels 1, 3, 5 and 7 of the multiresolution hierarchy reconstructed with full resolution of $512\times 512$, as well as the corresponding Fourier spectra (intermediate levels skipped to fit the page). 

The training times for each stage of the networks are as follows: 5s, 4s, 3s, 11s, 17s, 29s and 48s. The total training time is 117s. The machine was a Windows 10 laptop with a NVIDIA RTX A5000 Laptop GPU.
Note that these times result from the adaptive training regime and the number of samples for each level of the Gaussian Pyramid. 

\begin{figure}[!h]
\centering
\includegraphics[width=0.99\linewidth]{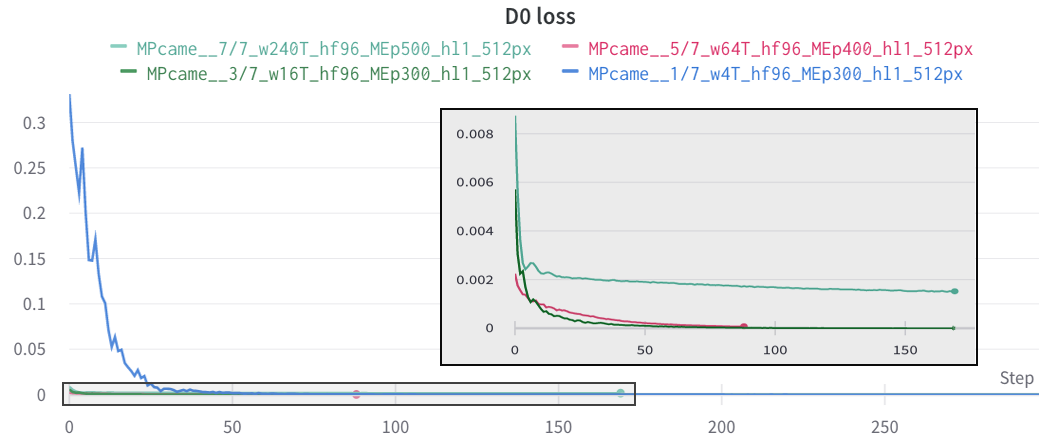}
\caption{Qualitative convergence behavior for Cameraman in Fig~\ref{f:lod}.}
\label{f:graph}
\end{figure}

The training evolution is depicted in the graph of Figure~\ref{f:graph} that shows the convergence of the MSE loss with the number of epochs for each multiresolution stages 1, 3, 5, and 7. It is worth pointing out the qualitative behavior of the network, in that the base level (stage 1) takes more than 200 epochs to reach the limit, while detail levels (stages 3, 5, 7) take less than 150 epochs to converge. Also, the error decreases for each level of detail. It is like, there are two different modes, one to fit the base level and other for the detail levels.

The inference time for image reconstruction, both in CPU and GPU is sufficiently fast for interactive visualization.

\subsection{Texture}

The second imaging example is the usage of the MR-Net representation to model texture and patterns. Arguably, visual textures constitute one of the most important applications for images in diverse fields, ranging from photo-realistic simulations to interactive games. Currently, more and more image rendering relies on some kind of graphics acceleration, sometimes through GPUs integrated with Neural Engines. In that context, it is desirable to have a neural image representation that is compact and supports a level of detail.

For the experiment shown in this subsection we have chosen an image of woven fabric background with patterns. 
The characteristics of this texture allows us to explore the limits of visual patterns at different resolutions. The original image has a resolution of $1025\times 1025$ pixels and the corresponding model contains 5 levels of detail.

\begin{figure}[!h]
\centering
\begin{subfigure}{.19\linewidth}
  \centering
\includegraphics[width=\linewidth]{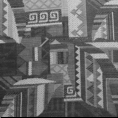}\\
\vspace{0.05cm}
\includegraphics[width=\linewidth]{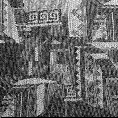}
\caption{}
\end{subfigure}
\begin{subfigure}{0.39\linewidth}
  \centering
\includegraphics[width=\linewidth]{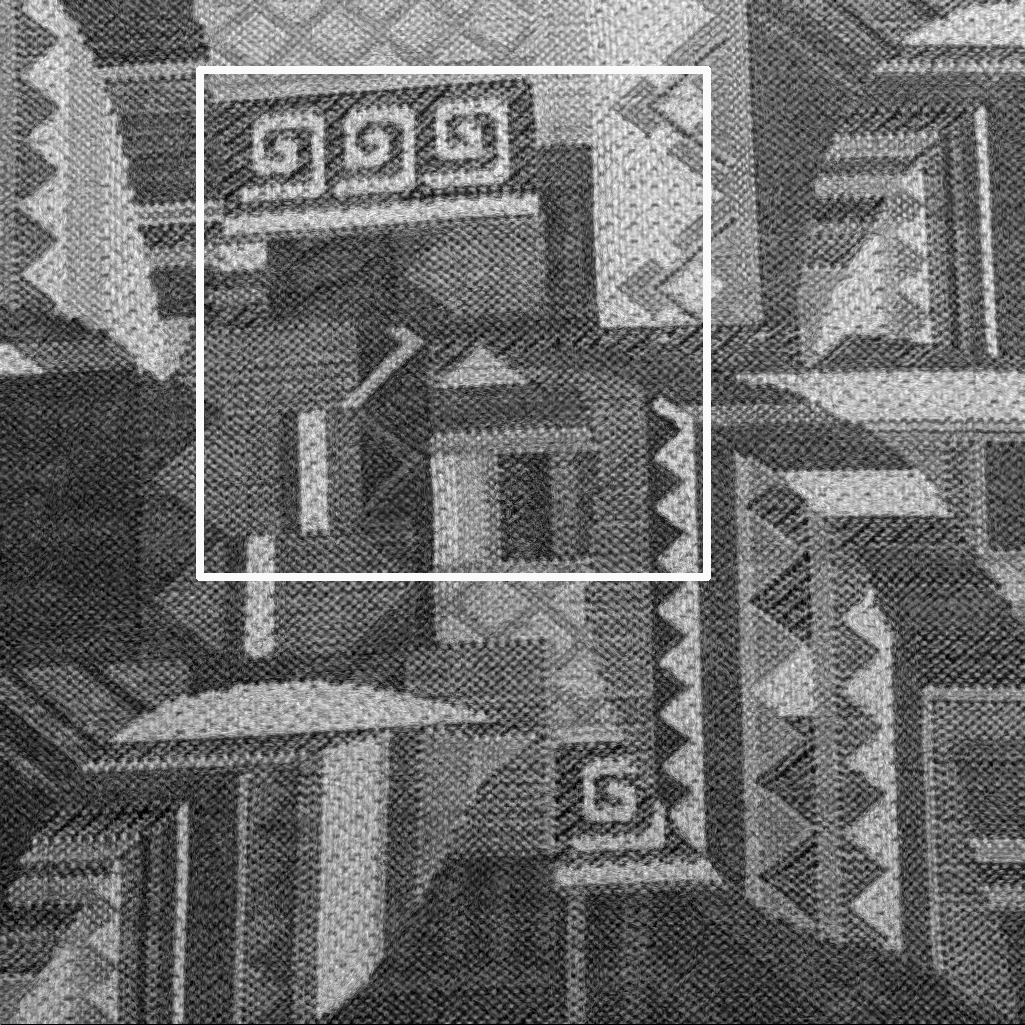}
\caption{}
\end{subfigure}
\begin{subfigure}{0.39\linewidth}
  \centering
\includegraphics[width=\linewidth]{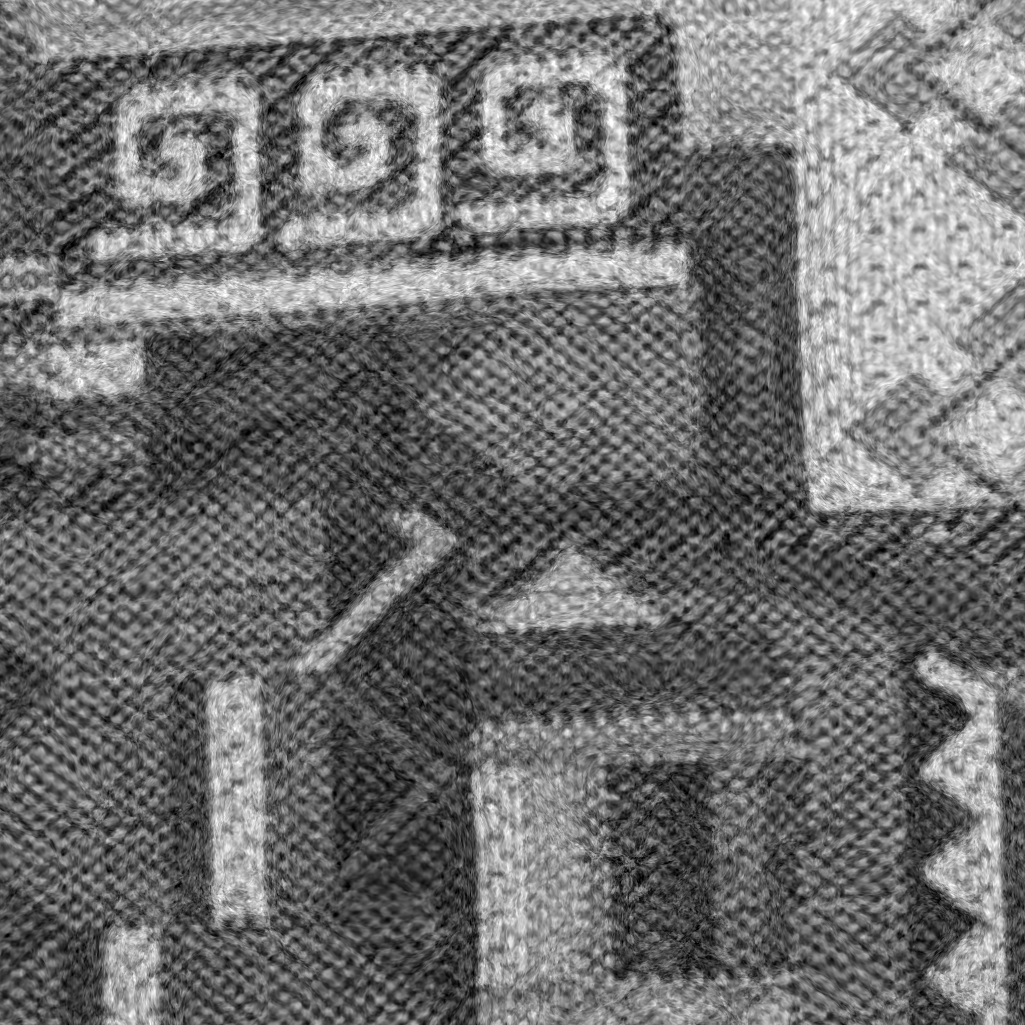}
\caption{}
\end{subfigure}
\caption{Woven Fabric Texture: (a) zoom out; (b) original image resolution; (c) zoom in.}
\label{f:pattern}
\end{figure}

In Figure~\ref{f:pattern} we show our reconstruction at original resolution, Fig.7(b) (center); as well as a zoom-in, Fig.7(c) (right), and a zoom-out, Fig.7(a) (left).

The zoom-in is a detail with resolution of $562\times 562$. marked by the white rectangle in Fig.7(b) and scaled up to $1025\times 1025$. Thus a zoom-in factor of 1.8 times. It can be seen that the enlargement extrapolates the fine details of the image at this higher resolution beyond the original image. 

The zoom-out is a reduction of the entire image to $118\times 118$ pixels (shown enlarged to $501\times 501$ in the image for better viewing). The top sub-image is the network reconstruction at the appropriate level of detail (approximately 0.92). The bottom sub-image is point-sample nearest neighbor reduction. It can be seen that our reconstruction is a proper anti-aliased rendition of the image, while the sampled reduction exhibits aliasing artifacts.

These two behaviors in the experiment are manifestations of ``magnification" and ``minification", classical resampling regimes for respectively scaling up and down the image~\cite{pixel}. In the first case it is necessary to interpolate the pixel values and in the second case it is required to integrate pixel values corresponding to the reconstructed pixel. The M-Net model accomplishes these tasks automatically.
Note that we have chosen a fractional scaling factors in both cases to demonstrate the continuous properties in space and scale of the M-Net~model.

\subsection{Anti-aliasing}

In the previous subsection we resorted to level of detail control to guarantee an alias free rendering independently of the sampling resolution. However, this task was facilitated because we could use a constant level of detail for the entire image, due to the zooming in/out operation in 2D.

On the other hand, in texture mapping applications, this scenario is no longer the case. Typically, it requires to map a 2D texture onto a 3D surface that is rendered in perspective by a virtual camera. In such situation, the level of detail varies spatially depending on the distance of the 3D surface point from the camera. Here, proper anti-aliasing must compensate the foreshortening caused by a projective transformation. Next we present a simple example of anti-aliasing using the M-Net.

Let $I$ be a checkerboard image, $T$ be a \textit{homography} mapping the pixel coordinates $(x,y)$ of the screen to the texture coordinates $(u,v)$ of $I$, and $f=g_1+\cdots+g_N:[-1,1]^2\to \mathbb{R}$ be a M-net with $N$ stages approximating $I$. 

\begin{figure}[!h]
\centering
\includegraphics[width=0.48\linewidth]{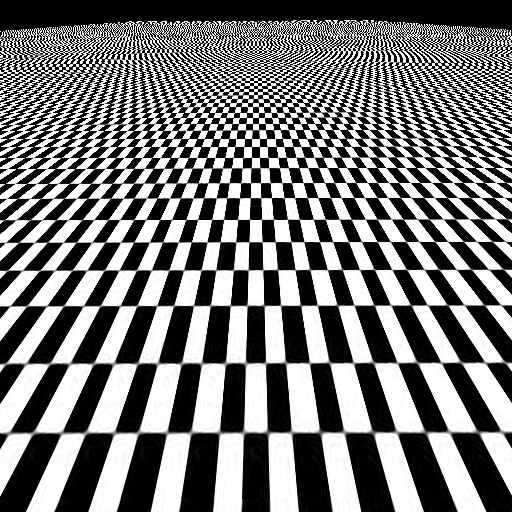}
\includegraphics[width=0.48\linewidth]{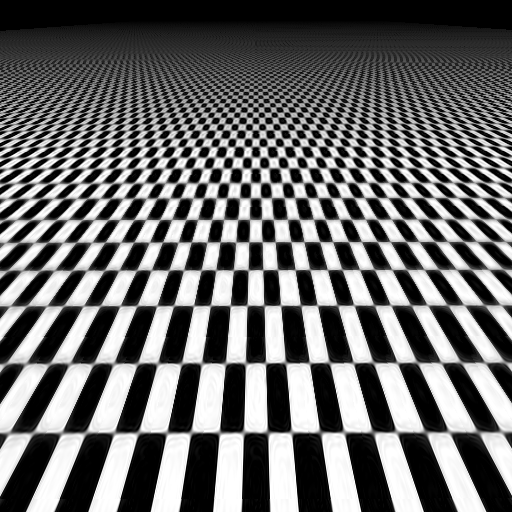}
\centerline{(a)\hfil\hfil(b)}
\caption{Checkerboard in perspective: (a) point sampled texture rendition; (b) M-Net anti-aliased reconstruction.}
\label{f:alias}
\end{figure}

Figure~\ref{f:alias}(left) illustrates aliasing effects on the image $I$ after applying it to the inverse of $T$.
We avoid such a problem using the multiresolution of the M-Net $f$. The result is presented in Figure~\ref{f:alias}(right). Observe that the procedure reduces aliasing at large distances. Specifically,
we define the \textit{level of detail parameter} $\lambda(x,y)$ for the M-Net $f$ at a pixel $(x,y)$ considering the formula proposed by Heckbert~\cite{heckbert1983texture}:
\begin{align*}
    \lambda(x,y)=\max\left\{\sqrt{\!\left(\frac{\partial u}{\partial x}\right)^2\!\!\!+\!\!\left(\frac{\partial v}{\partial x}\right)^2}\!, \sqrt{\!\left(\frac{\partial u}{\partial y}\right)^2\!\!\!+\!\!\left(\frac{\partial v}{\partial y}\right)^2}\right\}.
\end{align*}
Thus $\lambda(x,y)$ is the bigger length of the parallelogram generated by the vectors $\frac{\partial T}{\partial x}$ and $\frac{\partial T}{\partial y}$. 

We define the resolution $\lambda$ of $f$ using the formula
\begin{align*}
    f_\lambda=\lambda_1 g_1+\cdots+\lambda_Ng_N,
\end{align*}
where $\lambda_i$ are weights defined as follows.
First, scale $\lambda$ such that $\lambda\big([-1,1]^2\big)\subset[0,N]$. Thus, set $\lambda_i=1$ for $1\leq i\leq\lfloor \lambda \rfloor$, $\lambda_{\lfloor \lambda \rfloor}=\lambda- \lfloor \lambda \rfloor$, and $\lambda_i=0$ otherwise.
Here $\lfloor \lambda \rfloor$ denotes the \textit{floor} value of $\lambda$.


\section{Comparisons}

%
%
%

In this section we compare the performance of MR-Net image representation with two other neural network models, namely SIREN and BACON. For this evaluation we used the ``Cameraman" image shown in Subsection~\ref{ss:LOD}, comparing model size and reconstruction quality. The results are summarized in Table~\ref{t:comp}.

\begin{table}[!h]
\centering
\begin{tabular}{l|r|r|r}
\hline
Model & \# Params $\downarrow$ & PSNR $\uparrow$ & \# Levels \\
\hline
SIREN & 197K & 61.8 db & 1  \\
BACON & 398K & 82.1 db & 7 \\
MR-Net & {\bf 121K} & {\bf 84.9} db & 7  \\
\hline
\end{tabular}
\vspace{0.1cm}
\caption{\label{tab:comp} Comparison with SIREN and BACON.}
\label{t:comp}
\end{table}

The M-Net hyper-parameters are: 96 hidden features; $\omega_0 \in [4, 256]$ ; trained with a Gaussian Pyramid of 7 levels. Therefore, the model size has 121937 parameters. The image resolution is 512 x 512 pixels. The PSNR of the final image reconstruction is 84.9 db.

For SIREN we employed the configuration of the image experiments in their Github public code and paper. The network hyper-parameters are: 3 hidden layers; 256 hidden features; $\omega_0 = 120$. The model size has 197376 parameters and it has only 1 level of detail. The PSNR of the reconstructed image is 61.8 db.

For BACON we also based the configuration on their paper examples and public code at Github. However we made some adjustments to make the hyper-parameters compatible with SIREN and M-Net settings, as follows: image size $256\times 256$; 6 hidden layers; 256 hidden features. Accordingly, the total number of parameters is 398343. The PSNR of the level 6 image is 82.1 db. We remark that to establish a fair comparison with SIREN we evaluate only the PSNR of the final full resolution image.

Based on these experiments we conclude that MR-Net compared favorably in relation to SIREN and BACON, both in terms of representation size and quality of image reconstruction. Compared to SIREN the M-Net model is only 62 \% of the size of the SIREN model, even though it has 7 levels of detail in contrast with 1 level for SIREN. Additionally, the image reconstruction has 1.37 better quality, despite the fact that the representation is more compact.
Compared to BACON, the results are somewhat better. The M-Net model is just 30 \% of the size of BACON model, while our reconstruction of the final image has almost the same quality (i.e., 1.034 \%).


\subsection{Comparison with SIREN}

Parameters: 
Hidden Features =  256; 
Hidden Layers = 3; 
Number of Levels = 1; 
w0 parameter = 120; 
Model Size = 197376;
Image Resolution = 512 x 512.
The PSNR of Reconstruction = 61.8 db.

\begin{figure}[!h]
\centering
\includegraphics[width=0.4\linewidth]{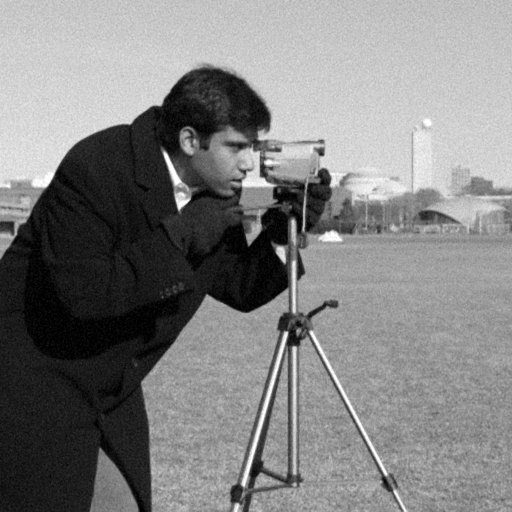}
\includegraphics[width=0.4\linewidth]{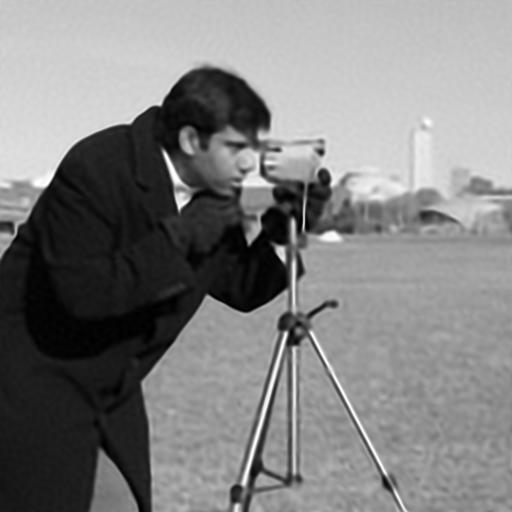} \\
\centerline{M-Net \hfil SIREN}
\caption{Comparison of reconstructed images.}
\label{f:siren}
\end{figure}

\noindent
Figure~\ref{f:siren} shows the image reconstructions with M-Net and SIREN models.

\newpage

\subsection{Comparison with BACON}

Parameters: 
Hidden Features = 256;
Hidden Layers = 6;
Learning Rate = 0.005;
Number of Steps = 5001;
Pe Scale = 3.0;
Model size = 398343.
The PSNR of Reconstruction =  82.1 db.

\begin{figure}[!h]
\centering
\includegraphics[width=0.93\linewidth]{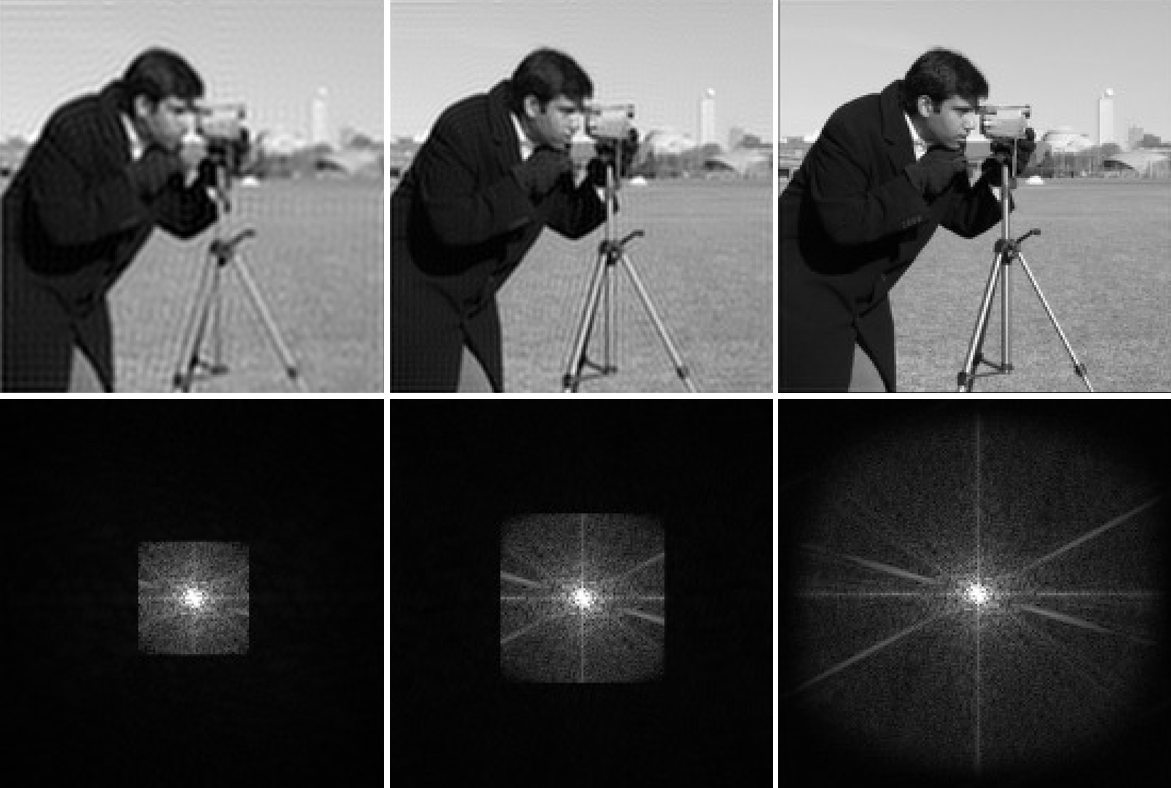}
\centerline{Level 1 \hfil Level 2 \hfil Level 6}
\caption{BACON image reconstruction and frequency spectra}
\label{f:bacon}
\end{figure}

\begin{figure}[!h]
\centering
\includegraphics[width=0.93\linewidth]{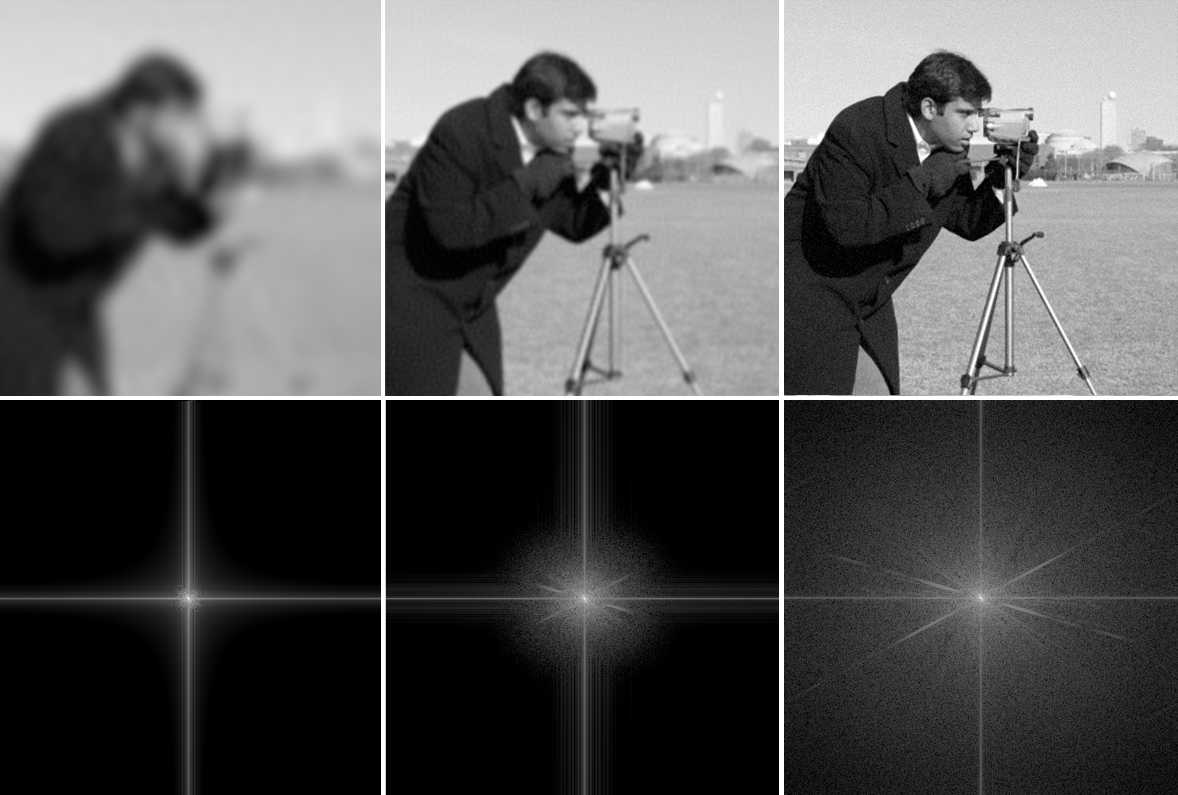}
\centerline{Level 3 \hfil Level 5 \hfil Level 7}
\caption{M-Net image reconstruction and frequency spectra}
\label{f:mnet}
\end{figure}

\noindent
Figures~\ref{f:bacon} and~\ref{f:mnet} show the image reconstructions and the corresponding frequency spectra for three levels of detail in the BACON and M-Net models respectively (1, 2, 6) and (3,5, 7). Note that we select these levels to better match the frequency spectra between the representations.

BACON controls the frequency band for Level of Detail by truncating the spectrum of the more detailed level (note that the center of the spectrum images for Levels 1 and 2 in Figure~\ref{f:bacon} are analogous to the center of the spectrum in Level 6). This approach is analogous to applying a low-pass filter with non-ideal shape in the frequency domain, which results in an image with ringing effect (notice how the silhouettes propagate all over the left image in Figure~\ref{f:bacon}). M-Net does not present such artifacts. Figure~\ref{f:mnet} resembles more faithfully what would be expected to be the process of sequentially applying Gaussian filters in a high resolution image.

\section{Considerations for Other MR-Net Variants}

In the paper, we described the MR-Net architecture and its three variants, namely: S-Net; L-Net and M-Net. however we have used only the M-Net variant for imaging applications in the paper. In the future we plan to explore the other two variants.

Nonetheless, here we it is appropriate to make a few considerations about the S-Net and L-Net variants.

The S-Net provides a neural image representation as a weighted sum of $\sin(x)$ functions. In that sense, S-Net is equivalent to BACON and other Multiplicative Filter Networks based on  sinusoidal atoms. As such, it is amenable to represent periodic visual patterns.

The L-Net image representation is composed of a sum of level-of-detail stages given by independent MR-Modules neural sub-networks. The relation of this representation with the Laplacian Pyramid makes it suitable for image operations in the gradient domain.

We intend study and compare in greater detail the three variants of the MR-Net architecture and also to develop imaging applications in directions pointed out above for S-Net and L-Net, complementing the ones we have presented in the paper for M-Net.

\section{Closing Remarks}

In this last section we close the paper with an assessment of our results, as well as, its limitations, and a discussion of future directions for our research.

\subsection{Limitations}

We found that choosing appropriate values for the $\omega_0$ parameter, which defines the spatial frequency of the first layer of the network, is important to achieve proper results. When using a shallow sinusoidal network such as the S-Net, we can use the Nyquist frequency as a direct reference to pick the frequency intervals at each stage. However, a deep sinusoidal network such as the M-Net can learn higher frequencies that were not present in the initialization. To the best of our knowledge, there is not (yet) theory to compute or bound these~frequencies. 

In our experiments, we determined the $\omega_0$ values for initialization empirically, testing values below the Nyquist frequency. To better harness the power of sinusoidal neural networks, it is important to develop mathematical theories to understand how the composition of sine functions introduces new frequencies based on the initialization of the network.

\subsection{Ongoing and Future Work}

In terms of future work, we plan to expand this research in two main directions. On one hand, we would like to explore the MR-Net architecture for other image applications including super-resolution, operations in the gradient domain, generation of periodic and quasi-periodic patterns, as well as image compression.
On the other hand, we would like to extend the MR-Net representation to other media signals in higher dimensions, such as video, volumes, and implicit surfaces.

\section*{Acknowledgment}

We thank Daniel Yukimura for  participating in this project.

\bibliographystyle{alpha}
\bibliography{main}

\end{document}